\documentclass[sigconf]{acmart}
\renewcommand\footnotetextcopyrightpermission[1]{}
\usepackage{xcolor}

\usepackage{microtype}
\usepackage{graphicx}
\usepackage{subfigure}
\usepackage{booktabs}
\usepackage{hyperref}

\usepackage{amsmath}
\usepackage{newtxmath}
\usepackage{amsthm}

\usepackage[capitalize,noabbrev]{cleveref}

\theoremstyle{plain}

\theoremstyle{definition}

\theoremstyle{remark}

\usepackage{multirow}

\newlength\savewidth
\newcommand\shline{\noalign{\global\savewidth\arrayrulewidth
                            \global\arrayrulewidth 1.5pt}%
                   \hline
                   \noalign{\global\arrayrulewidth\savewidth}
                   }

\usepackage{mathtools}
\DeclarePairedDelimiter\floor{\lfloor}{\rfloor}

\usepackage{algorithm}
\usepackage[frozencache,cachedir=.]{minted}

\usepackage{enumitem}
\setlist[itemize]{leftmargin=*, topsep=0.2em, itemsep=0pt, parsep=0pt, partopsep=0pt}
\setlist[enumerate]{leftmargin=*, topsep=0.2em, itemsep=0pt, parsep=0pt, partopsep=0pt}

\newlength{\enumerateparindent}
\newcommand{\proposed}{UltraSTF}

\setcopyright{acmlicensed}
\copyrightyear{2018}
\acmYear{2018}
\acmDOI{XXXXXXX.XXXXXXX}
\acmConference[Conference acronym 'XX]{Make sure to enter the correct
  conference title from your rights confirmation email}{June 03--05,
  2018}{Woodstock, NY}
\acmISBN{978-1-4503-XXXX-X/2018/06}

\begin{document}

\title[Ultra-Compact Model for Large-Scale Spatio-Temporal Forecasting]{\proposed{}: \underline{Ultra}-Compact Model for Large-Scale \underline{S}patio-\underline{T}emporal \underline{F}orecasting}

\author{Chin-Chia Michael Yeh}
\authornote{miyeh@visa.com}
\author{Xiran Fan}
\author{Zhimeng Jiang}
\author{Yujie Fan}
\affiliation{%
  \institution{Visa Research}
  \city{Foster City}
  \state{CA}
  \country{USA}
}

\author{Huiyuan Chen}
\author{Uday Singh Saini}
\author{Vivian Lai}
\author{Xin Dai}
\affiliation{%
  \institution{Visa Research}
  \city{Foster City}
  \state{CA}
  \country{USA}
}

\author{Junpeng Wang}
\author{Zhongfang Zhuang}
\author{Liang Wang}
\author{Yan Zheng}
\affiliation{%
  \institution{Visa Research}
  \city{Foster City}
  \state{CA}
  \country{USA}
}

\renewcommand{\shortauthors}{Chin-Chia Michael Yeh et al.}

\begin{abstract}
Spatio-temporal data, prevalent in real-world applications such as traffic monitoring, financial transactions, and ride-share demands, represents a specialized case of multivariate time series characterized by high dimensionality.
This high dimensionality necessitates computationally efficient models and benefits from applying univariate forecasting approaches through channel-independent strategies.
SparseTSF, a recently proposed competitive univariate forecasting model, leverages periodicity to achieve compactness by focusing on cross-period dynamics, extending the Pareto frontier in terms of model size and predictive performance.
However, it underperforms on spatio-temporal data due to limited capture of intra-period temporal dependencies.
To address this limitation, we propose \proposed{}, which integrates a cross-period forecasting component with an ultra-compact shape bank component.
Our model efficiently captures recurring patterns in time series using the attention mechanism of the shape bank component, significantly enhancing its capability to learn intra-period dynamics.
\proposed{} achieves state-of-the-art performance on the LargeST benchmark while utilizing fewer than 0.2\% of the parameters required by the second-best methods, thereby further extending the Pareto frontier of existing approaches.
The source code is available at: \url{https://sites.google.com/view/ultrastf}.
\end{abstract}


\keywords{Time Series, Forecasting, Ultra-Compact Model, Efficiency}


\maketitle

\section{Introduction}
Spatio-temporal data is ubiquitous in real-world applications, arising from activities such as traffic monitoring, transaction logging, and ride-share demands~\cite{liu2023largest}.
This data type exhibits defining characteristics that distinguish it from general multivariate time series: high dimensionality.
For instance, the highest-dimensional dataset in the LargeST~\cite{liu2023largest} spatio-temporal benchmark contains 8,600 dimensions, far exceeding the 862 dimensions found in popular long-term time series forecasting (LTSF) datasets~\cite{zeng2023transformers}.
Such scale requires computationally efficient approaches for practical applications.
The channel-independent strategy~\cite{han2024capacity}, where a univariate forecasting model is applied independently to each dimension of the data, becomes increasingly attractive for these high-dimensional scenarios.

Recently, lightweight models like SparseTSF~\cite{lin2024sparsetsf} have shown competitive performance against the state-of-the-art time series forecasting methods.
Utilizing fewer than 1K parameters, SparseTSF achieves performance comparable to models with over a million parameters.
This efficiency greatly reduces computational costs and extends the Pareto frontier with respect to both model size and prediction errors, making SparseTSF attractive for deployment.

\begin{figure}[htp]
\begin{center}
\centerline{
\includegraphics[width=0.99\linewidth]{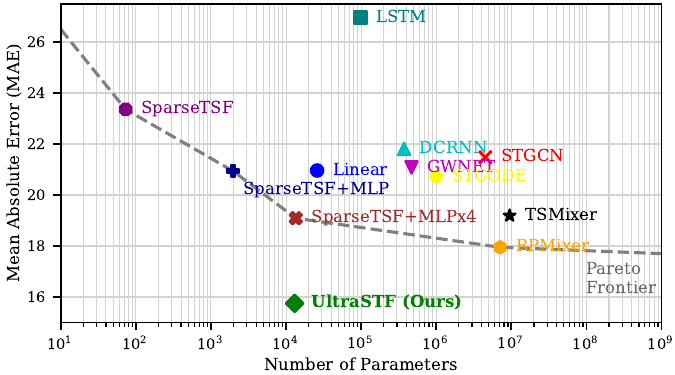}
}
\caption{
The proposed method extends the Pareto frontier established by existing methods on the CA dataset.
}
\label{fig:n_param_error}
\end{center}
\end{figure}

\begin{figure}[htp]
\begin{center}
\centerline{
\includegraphics[width=0.99\linewidth]{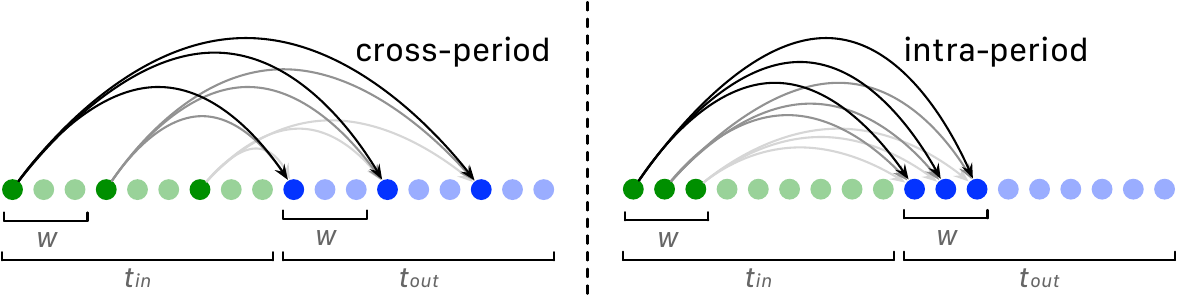}
}
\caption{
The difference between cross-period relationships and intra-period relationships.
}
\label{fig:inter_intra}
\end{center}
\end{figure}

However, applying SparseTSF to spatio-temporal data via the channel independent strategy reveals notable performance limitations, as illustrated in \cref{fig:n_param_error}.
SparseTSF and its variants underperform on such data, likely due to limited expressive capabilities.
Specifically, SparseTSF's compactness stems from segmenting the input time series into subsequences based on a user-specified period~$w$, focusing solely on modeling cross-period relationships (\cref{fig:inter_intra}.\textit{left}) while neglecting intra-period temporal dependencies (\cref{fig:inter_intra}.\textit{right}).
As shown in \cref{fig:inter_intra}, since the cross-period model is shared across different time steps within a period, if the cross-period relationship changes over time within a period, modeling solely cross-period relationships will not suffice to capture the overall temporal dynamics.
In such cases, incorporating intra-period relationships becomes crucial for accurate forecasting.

In this paper, we introduce \proposed{}, a compact and efficient spatio-temporal forecasting method that addresses the limitations of previous approaches.
\proposed{} integrates a cross-period sparse forecasting component with a novel ultra-compact shape bank component to model both cross-period and intra-period relationships.
The shape bank learns a set of patterns of length~$w$, each represented as a key-value pair, and employs an attention mechanism to compute the output based on the input subsequence (or \textit{query}).
This lightweight attention module is applied independently to each subsequence, focusing on modeling intra-period dynamics.
By leveraging the channel independent strategy and combining these components, \proposed{} achieves superior performance.
As demonstrated in \cref{fig:n_param_error}, \proposed{} delivers state-of-the-art accuracy while maintaining a minimal model size, outperforming all comparable methods with fewer parameters and extending the Pareto frontier established by existing methods.
Our contributions are:

\begin{itemize}
    \item We identify the importance of modeling both cross-period and intra-period relationships in compact spatio-temporal forecasting models, addressing a significant limitation in existing approaches.
    \item We propose \proposed{}, a novel method that integrates cross-period sparse forecasting with an ultra-compact shape bank mechanism, effectively capturing complex temporal dynamics while maintaining an extremely small parameter count.
    \item Through experiments across diverse datasets, we demonstrate that \proposed{} achieves state-of-the-art performance using only 0.2\% of the parameters required by previous leading methods, substantially advancing the efficiency-accuracy trade-off in spatio-temporal forecasting.
\end{itemize}

\section{Related Work}
\label{sec:related}
Spatio-temporal forecasting methods typically integrate two key components:
1) graph modeling architectures, such as graph convolutional networks (GCNs)~\cite{yu2017spatio} or graph attention networks (GATs)~\cite{guo2019attention}, and
2) sequential modeling architectures like recurrent neural networks (RNNs)~\cite{li2017diffusion}, convolutional networks~\cite{wu2019graph}, or transformers~\cite{vaswani2017attention}.
The graph modeling components process spatial relationships from input graph data, while sequential modeling components capture temporal dependencies in time series data.

Recent models in this domain exemplify this integration of spatial and temporal modeling, including DCRNN~\cite{li2017diffusion}, STGCN~\cite{yu2017spatio}, ASTGCN~\cite{guo2019attention}, GWNET~\cite{wu2019graph}, STGODE~\cite{fang2021spatial}, D$^2$STGNN~\cite{shao2022decoupled}, and DGCRN~\cite{li2023dynamic}.
An alternative approach, explored by methods such as AGCRN~\cite{bai2020adaptive}, DSTAGNN~\cite{lan2022dstagnn}, and RPMixer~\cite{yeh2024rpmixer}, bypasses reliance on predefined spatial graphs by instead inferring relationships between locations directly from time series data.
Despite their differences, many of these approaches learn dependencies between different locations, resulting in model sizes of at least $O(n)$ with respect to the number of locations.

Recent research has highlighted the competitive performance of linear models employing an independent channel strategy, particularly when compared to more complex transformer-based solutions~\cite{zeng2023transformers}.
Compact architectures, such as SparseTSF~\cite{lin2024sparsetsf}, further push the boundaries of efficiency while maintaining strong predictive performance.
Building on this thread, \proposed{} aims to advance the compactness frontier for spatio-temporal forecasting models while preserving or enhancing predictive capabilities.

\section{Preliminary}
\label{sec:preliminary}
\begin{figure}[htp]
\begin{center}
\centerline{
\includegraphics[width=0.99\linewidth]{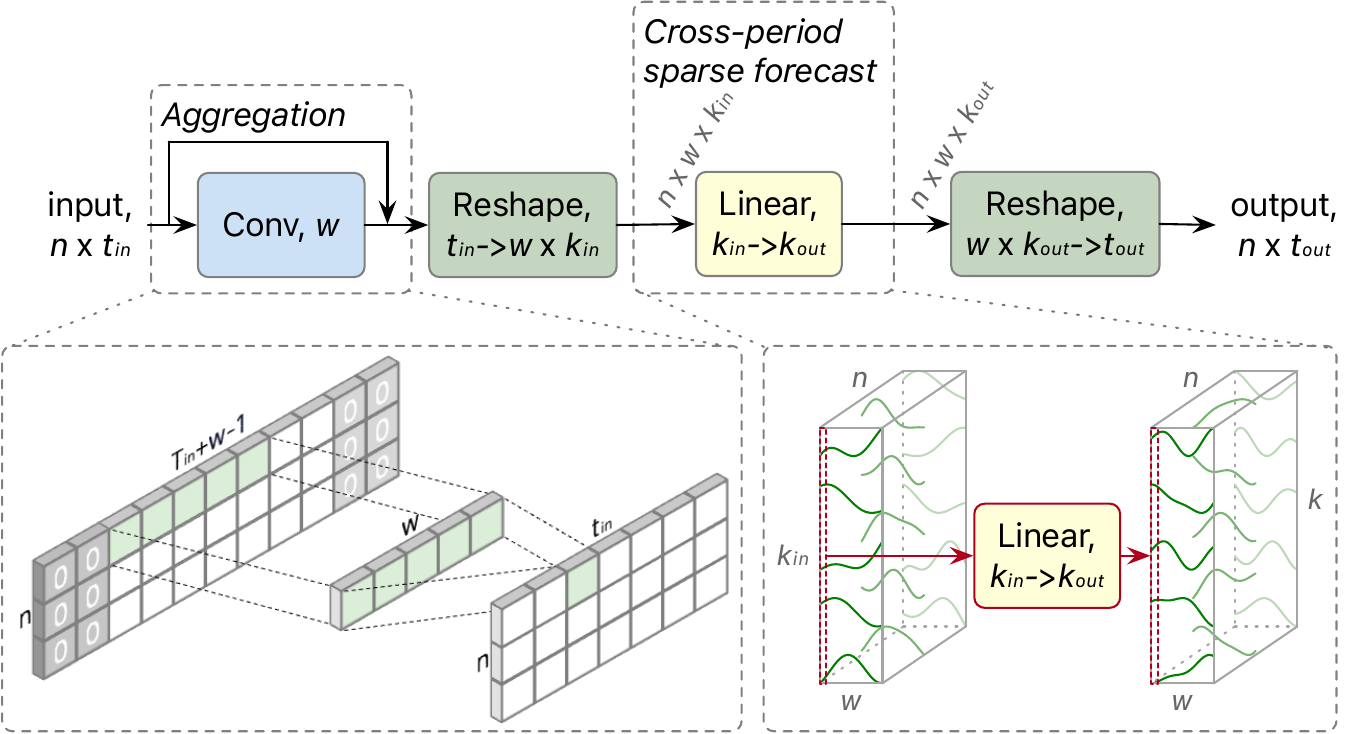}
}
\caption{
The design of the SparseTSF model.
We use $n$, $t_\text{in}$, $w$, and $t_\text{out}$ to denote the number of spatial locations, input time steps, user-specified period, and output time steps, respectively.
Note that $t_\text{in} \coloneq k_\text{in} \times w$ and $t_\text{out} \coloneq k_\text{out} \times w$.
}
\label{fig:sparsetsf}
\end{center}
\end{figure}

This section provides an overview of the SparseTSF method and motivates our proposed approach.
As illustrated in \cref{fig:sparsetsf}, SparseTSF processes input time series through two main components:
1) An aggregation component that computes a weighted sum over a specified period length~$w$ using a convolutional layer.
This component reshapes the input time series of length~$t_\text{in}$ into $k_\text{in}$ non-overlapping subsequences of length~$w$, where $t_\text{in} = k_\text{in} \times w$.
2) A cross-period sparse forecast component that models temporal relationships between corresponding time steps in the input and output subsequences.
While effective, SparseTSF's approach has limitations, notably its lack of explicit modeling of intra-subsequence temporal dependencies beyond basic aggregation.
This limitation presents an opportunity for enhancement that \proposed{} addresses.

To verify the importance of simultaneously modeling both cross-period and intra-period relationships in spatio-temporal forecasting, we conducted a comparative analysis.
We replaced the SparseTSF model with a more comprehensive fully-connected linear model.
\cref{tab:preliminary} summarizes the performance and model size comparison, with detailed experimental setup described in \cref{sec:experiment}.
The results demonstrate that the linear model outperforms SparseTSF, underscoring the significance of capturing both cross-period and intra-period relationships.
However, the fully-connected layer's significantly larger model size is a major drawback.
To address this trade-off between performance and efficiency, we propose the \proposed{} model, which effectively balances both aspects.

\begin{table}[htp]
\caption{
Linear model outperforms SparseTSF at the cost of approximately 300 times larger model size.
Performances are reported as average Mean Absolute Error (MAE).
}
\label{tab:preliminary}
\begin{center}
\resizebox{0.75\linewidth}{!}{%
\begin{tabular}{lccccc}
    \shline
    \multirow{2}{*}{Method} & \multirow{2}{*}{Param} & \multicolumn{3}{c}{Data} \\ \cline{3-6} 
      &  & SD & GBA & GLA & CA \\ 
     \hline \hline
     Linear & 26K & \textbf{23.01} & \textbf{22.68} & \textbf{22.85} & \textbf{20.96} \\
     SparseTSF & 73 & 24.66 & 25.55 & 25.06 & 23.36 \\
     \shline
\end{tabular}%
}
\end{center}
\vskip -0.1in
\end{table}

Notably, our finding that the linear model outperforms SparseTSF contrasts with the results on LTSF datasets presented in~\cite{lin2024sparsetsf}.
This discrepancy highlights the increased importance of capturing both cross-period and intra-period relationships in spatio-temporal datasets compared to LTSF datasets.
Consequently, we focus our experimental efforts on spatio-temporal datasets.

\section{Methodology}
\label{sec:method}
\begin{figure}[htp]
\begin{center}
\centerline{
\includegraphics[width=0.99\linewidth]{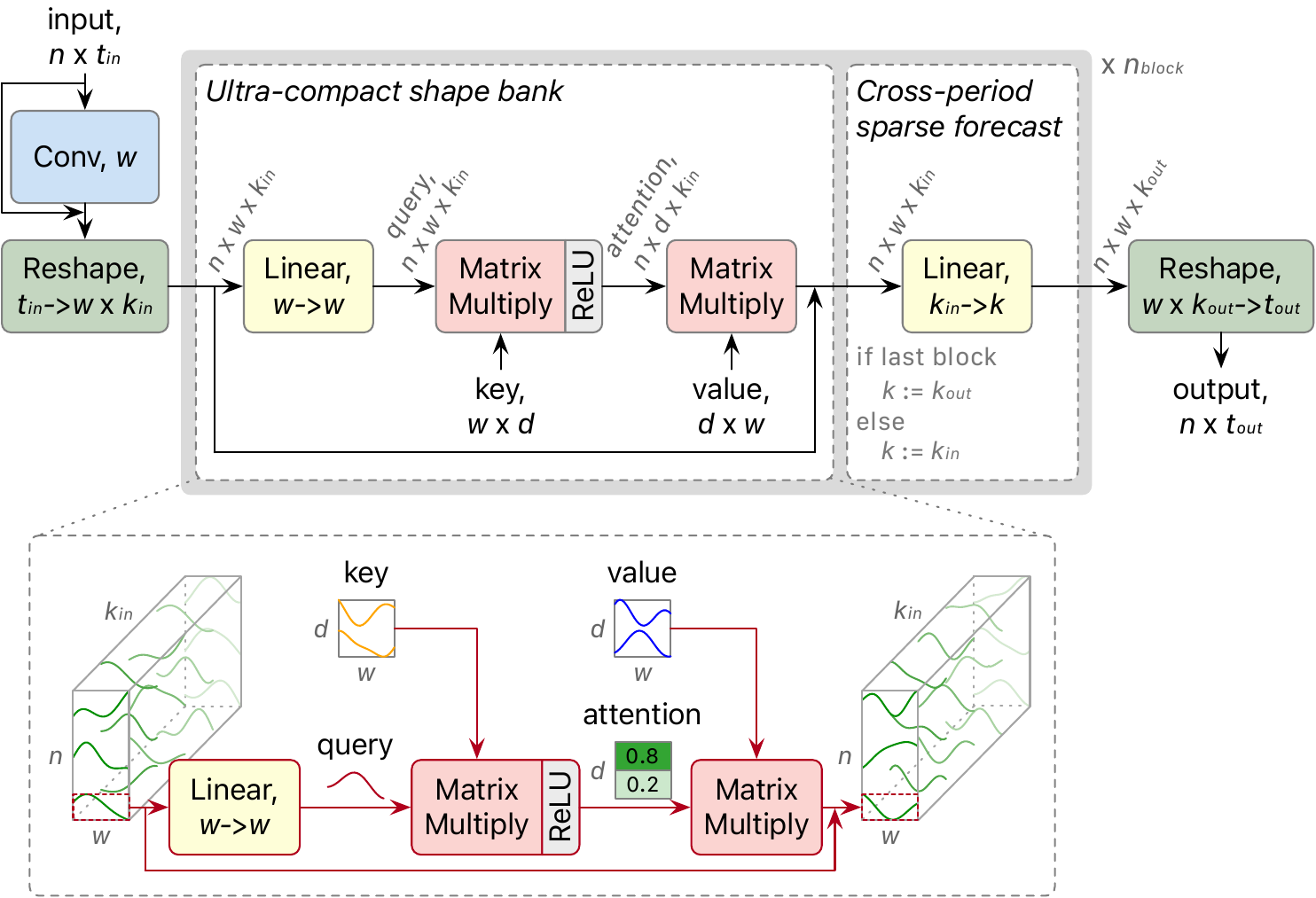}
}
\caption{
The design of the \proposed{} model.
We use $n$, $t_\text{in}$, $w$, $d$, and $t_\text{out}$ to denote the number of spatial locations, input time steps, user-specified period, number of learned shapes, and output time steps, respectively.
Note that $t_\text{in} \coloneq k_\text{in} \times w$ and $t_\text{out} \coloneq k_\text{out} \times w$.
}
\label{fig:overall}
\end{center}
\end{figure}

The overall model design of the proposed \proposed{} is shown in \cref{fig:overall}.
Given a spatio-temporal time series with length~$t_\text{in}$ and $n$ dimensions/spatial locations, the input time series is first processed with the convolutional/aggregation module.
The aggregation module computes the weighted sum along the time dimension at each time step for each spatial location.
This module can be implemented with a convolutional layer with a filter size of $w$ and a skip connection where $w$ is the user-specified period.
The bottom-left insert of \cref{fig:sparsetsf} illustrates the computation of the weighted sum using the convolutional layer.
There are two reasons behind the aggregation design.
First, it prevents the loss of information, as each time step after the aggregation module now has the contextual information from other time steps within the same period.
Second, the aggregation module mitigates the impact of outliers.

After the aggregation module, the input time series is reshaped from $n \times t_\text{in}$ to $n \times w \times k_\text{in}$ where $t_\text{in} \coloneq w \times k_\text{in}$.
This operation prepares the time series for the core blocks by segmenting each dimension of the input time series into $k_\text{in}$ subsequences of length $w$ (i.e., the period).
Segmenting the time series into chunks based on $w$ is necessary because the core blocks explicitly model both cross-period and intra-period relationships.
The reshaped time series is then fed into a sequence of core blocks.
Within each block, the input time series is first processed by the shape bank component and subsequently by the cross-period sparse forecast component.
Details of the cross-period sparse forecast component and the shape bank component are provided in \cref{sec:sparseforecast,sec:shapebank}, respectively.

\subsection{Cross-Period Sparse Forecast}
\label{sec:sparseforecast}
We utilize the periodicity assumption to analyze the effectiveness of the cross-period sparse forecast component.
Given a univariate time series~$X$ and its period $w$, $X$ can be decomposed into the periodic component $P$ and trend component $T$ where

\begin{gather}
X = P + T \label{eq:decompose} \\[1ex]
P[i] = P[i + w] \label{eq:period}
\end{gather}

Assume the length of $X$ is $t_\text{in} + t_\text{out}$ where the first $t_\text{in}$ time steps are the input and the last $t_\text{out}$ time steps are the prediction target.
Given that $k_\text{in} = \frac{t_\text{in}}{w}$, $k_\text{out} = \frac{t_\text{out}}{w}$, $X_\text{in}=X[0:t_\text{in}]$, and $X_\text{out}=X[t_\text{in}:t_\text{out}]$, for any $i \in [0, \ldots, k_\text{out}]$ and $l \in [0, \ldots, w]$ the cross-period sparse forecast component can be formulated with \cref{eq:sparse_forecast_0} where $f_i$ is the sparse forecast function for the $i$th output subsequence.

\begin{equation}
\begin{aligned}
\label{eq:sparse_forecast_0}
X_\text{out}[iw+l] = f_i(&X_\text{in}[l], X_\text{in}[w+l], \\
&\ldots, X_\text{in}[(k_\text{in} - 1)w+l])
\end{aligned}
\end{equation}

\noindent For simplicity, we denote \cref{eq:sparse_forecast_0} as

\begin{equation}
\label{eq:sparse_forecast_1}
X_\text{out}[iw+l] = f_i(X_\text{in}[jw+l]), j = 0, \ldots, k_\text{in}.
\end{equation}

\noindent If we plug \cref{eq:decompose} into \cref{eq:sparse_forecast_1}, we get \cref{eq:sparse_forecast_2}.

\begin{equation}
\label{eq:sparse_forecast_2}
P_\text{out}[iw+l] + T_\text{out}[iw+l] = f_i(P_\text{in}[jw+l] + T_\text{in}[jw+l])
\end{equation}

\noindent Due to the periodicity described in \cref{eq:period}, \cref{eq:sparse_forecast_2} can be simplified to \cref{eq:sparse_forecast_3}.

\begin{equation}
\label{eq:sparse_forecast_3}
P[l] + T_\text{out}[iw+l] = f_i(P[l] + T_\text{in}[jw+l])
\end{equation}

\noindent Given that $P[l]$ is a fixed value for a given $l$ and is present in both the input and output of the forecasting function, the parameters of the cross-period sparse forecast component focus on learning the trend of the time series.

This analysis highlights both the type of information captured by the cross-period sparse forecast component and the information that is \textit{not} captured by the component.
Since the function~$f_i$ is fixed regardless of $l$, it cannot learn the correct relationship between $T_\text{out}$ and $T_\text{in}$ when the mapping between them varies with respect to $l$, regardless of the architecture used for $f_i$.
The ultra-compact shape bank component is proposed to address this issue.

\subsection{Ultra-Compact Shape Bank}
\label{sec:shapebank}
The proposed ultra-compact shape bank component is shown in \cref{fig:overall}.\textit{bottom}.
This component operates independently on each subsequence, with weights shared across all subsequences.
For a given subsequence, we first generate a query using a linear layer.
Assuming the user has specified $d$ shapes in the shape bank, we multiply the query with the key associated with each shape to obtain attention scores.
Following \cite{shen2023study}, we employ the ReLU activation function instead of softmax when computing the attention scores for improved efficiency.
Finally, we multiply the attention scores by the corresponding shapes (or \textit{values}) in the shape bank to produce the output.

We name this mechanism a ``shape bank" because it learns a collection of shapes that serve as the basis for the output time series.
Our shape bank design is considered ultra-compact as the entire component processes only one period (i.e., a size $w$ vector) rather than the whole time series (i.e., a size $w \times k_\text{in}$ vector).
This design significantly reduces computational complexity while maintaining model expressiveness.

We can use the same univariate time series $X$ to analyze the effectiveness of the shape bank component.
This time, we use the function~$f$ to represent the proposed shape bank component.
For any $i \in [0, w, 2w, \ldots, k_\text{out}w]$, the output is computed with \cref{eq:light_attn_0}.

\begin{equation}
\label{eq:light_attn_0}
X_\text{out}[i:i+w] = f(X_\text{in}[i-t_\text{in}:i-t_\text{in}+w])
\end{equation}

\noindent If we also plug \cref{eq:decompose} into \cref{eq:light_attn_0}, we get \cref{eq:light_attn_1}.

\begin{equation}
\begin{aligned}
\label{eq:light_attn_1}
P_\text{out}[i:i+w] + T_\text{out}[i:i+w] = \hphantom{XXXXX}\\
\begin{aligned}
f(&P_\text{in}[i-t_\text{in}:i-t_\text{in}+w]+  \\ 
&T_\text{in}[i-t_\text{in}:i-t_\text{in}+w])
\end{aligned}
\end{aligned}
\end{equation}

\noindent Given that $t_\text{in}$ is a multiple of $w$, we can simplify \cref{eq:light_attn_1} into \cref{eq:light_attn_2} with \cref{eq:period}.

\begin{equation}
\begin{split}
\label{eq:light_attn_2}
P[0:w] + T_\text{out}[i:i+w] = \hphantom{XXXXXXXX}\\
f(P[0:w] + T_\text{in}[i-t_\text{in}:i-t_\text{in}+w])
\end{split}
\end{equation}

\noindent Since $P[0:w]$ is a fixed vector in both the input and output of $f$, the shape bank component is once again focused on modeling the trend of the time series.
However, by comparing \cref{eq:light_attn_2} with \cref{eq:sparse_forecast_3}, we can observe that the shape bank component focuses on modeling the trend variation within a period, while the sparse forecasting component focuses on the cross-period trend. 
Together, the shape bank component and the sparse forecasting component complement each other.
For implementation details of \proposed{}, including multi-head extension, see \cref{app:implementation}.

\subsection{Complexity Analysis}
\label{sec:complexity}
We analyze the complexity of our proposed method in terms of time complexity, space complexity, and number of model parameters.
\Cref{tab:complexity} summarizes the results of our complexity analysis for the proposed \proposed{} method, alongside the complexities of the lightest variant of SparseTSF --- our baseline method with the lowest computational requirements.
Notably, despite the superior performance demonstrated in \cref{fig:n_param_error}, \proposed{} maintains identical time and space complexities to the lightest variant of SparseTSF.
The only difference lies in a marginally higher parameter count for \proposed{}, which accounts for the additional shape bank and multi-layer design.
For a comprehensive understanding, we provide a detailed complexity analysis in \cref{app:complexity}.
 
\begin{table}[htp]
\caption{Complexity analysis. Param: model size.}
\label{tab:complexity}
\begin{center}
\resizebox{\linewidth}{!}{%
\begin{tabular}{l||c|c}
\shline
 & \proposed{} & SparseTSF \\ \hline \hline
Time & $O(n t_\text{in}^2)$ & $O(n t_\text{in}^2)$ \\ 
Space & $O(n t_\text{in})$ & $O(n t_\text{in})$ \\ 
Param & 
{$\!\begin{aligned}
\floor*{\frac{w}{2}} + 1 & + (n_\text{block} - 1) \left(w^2 + 2wd + \frac{t_\text{in}^2}{w^2} \right) \\ 
& + \left(w^2 + 2wd + \frac{t_\text{in} t_\text{out}}{w^2} \right) \\
\end{aligned}$}
& 
$\floor*{\frac{w}{2}} + 1 + \left(\frac{t_\text{in} t_\text{out}}{w^2} \right)$  \\ \shline
\end{tabular}%
}
\end{center}
\vskip -0.1in
\end{table}

\section{Experiment}
\label{sec:experiment}
This section presents our experimental setup and results.
We describe datasets and baselines (\cref{sec:dataset,sec:baseline}), analyze benchmark results (\cref{sec:benchmark}), evaluate generalizability and scaling (\cref{sec:generalization,sec:scaling}), study \proposed{}'s hyperparameters (\cref{sec:hyper_parameter}), provide visualizations (\cref{sec:visualization}) and perform sensitivity analysis (\cref{sec:sensitivity}).

\subsection{Dataset and Benchmark Setting}
\label{sec:dataset}
Our experiments utilized the LargeST datasets~\cite{liu2023largest}, which encompass traffic data from 8,600 sensors across California.
Following the methodology described in~\cite{liu2023largest}, we generated four distinct sub-datasets: SD, GBA, GLA, and CA.
These sub-datasets represent sensor data from San Diego (716 sensors), the Greater Bay Area (2,352 sensors), Greater Los Angeles (3,834 sensors), and the entire California, respectively.
To ensure consistency with the experimental setup in~\cite{liu2023largest}, we focused exclusively on traffic data from the year 2019.
The original sensor readings, recorded every 5 minutes, were aggregated into 15-minute intervals, resulting in 96 intervals per day and a total of 35,040 time steps.
Each sub-dataset was then split into training, validation, and test sets in a 6:2:2 ratio.
The primary task was to forecast the next 12 intervals for each sensor at each timestamp.
We evaluated performance using Mean Absolute Error (MAE), Root Mean Squared Error (RMSE), and Mean Absolute Percentage Error (MAPE).

\subsection{Baseline Method}
\label{sec:baseline}
Our benchmark experiments involve 19 baseline methods.
First, we include the 11 baseline methods evaluated by \cite{liu2023largest}: HL, LSTM~\cite{hochreiter1997long}, DCRNN~\cite{li2017diffusion}, STGCN~\cite{yu2017spatio}, ASTGCN~\cite{guo2019attention}, GWNET~\cite{wu2019graph}, AGCRN~\cite{bai2020adaptive}, STGODE~\cite{fang2021spatial}, DSTAGNN~\cite{lan2022dstagnn}, D$^2$STGNN~\cite{shao2022decoupled}, and DGCRN~\cite{li2023dynamic}.
We also include SparseTSF~\cite{lin2024sparsetsf}, a very compact forecasting model.
To address the limited learning capabilities of SparseTSF, we consider two variants with increased parameter counts: SparseTSF+MLP, which replaces the linear model with a two-layer multilayer perceptron (MLP) as suggested by~\cite{lin2024sparsetsf}, and SparseTSF+MLP$\times 4$, which uses an eight-layer MLP instead of a two-layer MLP.
To demonstrate the benefits of our proposed ultra-compact attention module, we introduce SparseTSF+SA, a variant of \proposed{} that uses standard attention for the shape bank mechanism.
Additionally, we experiment with other general time series forecasting methods, including the linear model, TSMixer~\cite{chen2023tsmixer}, RPMixer~\cite{yeh2024rpmixer}, and iTransformer~\cite{liu2023itransformer}.

\begin{table*}[htp]
\caption{Performance comparisons. 
We \textbf{bold} the best-performing results and \underline{underline} the second-best results.
The performance reported in the ``Average" column is computed by averaging over 12 predicted time steps.
The absence of baselines on the datasets indicates that the models incur an out-of-memory issue.
Param: model size. K: $10^3$. M: $10^6$.}
\label{tab:largest_benchmark}
\begin{center}
\resizebox*{!}{0.93\textheight}{%
\begin{tabular}{llcccc|ccc|ccc|ccc}
    \shline
    \multirow{2}{*}{Data} & \multirow{2}{*}{Method} & \multirow{2}{*}{Param} & \multicolumn{3}{c}{Horizon 3} & \multicolumn{3}{c}{Horizon 6} & \multicolumn{3}{c}{Horizon 12} & \multicolumn{3}{c}{Average} \\ \cline{4-15} 
     &  &  & MAE & RMSE & MAPE & MAE & RMSE & MAPE & MAE & RMSE & MAPE & MAE & RMSE & MAPE \\ 
     \hline \hline
     \multirow{20}{*}{SD} & HL & -- & 33.61 & 50.97 & 20.77\% & 57.80 & 84.92 & 37.73\% & 101.74 & 140.14 & 76.84\% & 60.79 & 87.40 & 41.88\% \\
     & Linear & 26K & 19.27 & 31.88 & 14.67\% & 23.76 & 39.41 & 17.06\% & 25.13 & 44.60 & 18.01\% & 23.01 & 38.43 & 17.03\% \\
     & LSTM & 98K & 19.17 & 30.75 & 11.85\% & 26.11 & 41.28 & 16.53\% & 38.06 & 59.63 & 25.07\% & 26.73 & 42.14 & 17.17\% \\
     & ASTGCN & 2.2M & 20.09 & 32.13 & 13.61\% & 25.58 & 40.41 & 17.44\% & 32.86 & 52.05 & 26.00\% & 25.10 & 39.91 & 18.05\% \\
     & DCRNN & 373K & 17.01 & 27.33 & 10.96\% & 20.80 & 33.03 & 13.72\% & 26.77 & 42.49 & 18.57\% & 20.86 & 33.13 & 13.94\% \\
     & AGCRN & 761K & 16.05 & 28.78 & 11.74\% & 18.37 & 32.44 & 13.37\% & 22.12 & 40.37 & 16.63\% & 18.43 & 32.97 & 13.51\% \\
     & STGCN & 508K & 18.23 & 30.60 & 13.75\% & 20.34 & 34.42 & 15.10\% & 23.56 & 41.70 & 17.08\% & 20.35 & 34.70 & 15.13\% \\
     & GWNET & 311K & 15.49 & \underline{25.45} & \underline{9.90\%} & 18.17 & 30.16 & 11.98\% & 22.18 & 37.82 & 15.41\% & 18.12 & 30.21 & 12.08\% \\
     & STGODE & 729K & 16.76 & 27.26 & 10.95\% & 19.79 & 32.91 & 13.18\% & 23.60 & 41.32 & 16.60\% & 19.52 & 32.76 & 13.22\% \\
     & DSTAGNN & 3.9M & 17.83 & 28.60 & 11.08\% & 21.95 & 35.37 & 14.55\% & 26.83 & 46.39 & 19.62\% & 21.52 & 35.67 & 14.52\% \\
     & DGCRN & 243K & 15.24 & 25.46 & 10.09\% & 17.66 & 29.65 & \underline{11.77\%} & \underline{21.38} & 36.67 & 14.75\% & 17.65 & 29.70 & \underline{11.89\%} \\
     & D$^2$STGNN & 406K & \underline{14.85} & \textbf{24.95} & 9.91\% & \underline{17.28} & \underline{29.05} & 12.17\% & 21.59 & 35.55 & 16.88\% & \underline{17.38} & \underline{28.92} & 12.43\% \\
     & TSMixer & 1.4M & 15.92 & 25.72 & 10.73\% & 17.94 & 29.16 & 12.53\% & 21.45 & \underline{35.34} & 14.81\% & 18.10 & 29.40 & 12.33\% \\ 
     & RPMixer & 830K & 16.60 & 26.74 & 11.31\% & 19.07 & 30.77 & 13.80\% & 21.48 & 36.16 & 14.93\% & 18.73 & 30.57 & 13.14\% \\ 
     & iTransformer & 6.5M & 17.10 & 27.35 & 11.94\% & 19.31 & 31.86 & 12.33\% & 23.74 & 39.20 & 16.25\% & 19.63 & 32.02 & 13.14\% \\
     \cline{2-15}
     & SparseTSF & 73 & 24.51 & 46.34 & 15.92\% & 24.53 & 46.34 & 15.93\% & 25.72 & 47.34 & 18.40\% & 24.66 & 46.47 & 16.22\% \\
     & SparseTSF+MLP & 2K & 21.73 & 40.24 & 14.63\% & 21.74 & 40.22 & 14.62\% & 22.58 & 41.21 & 15.68\% & 21.85 & 40.38 & 14.77\% \\
     & SparseTSF+MLP$\times 4$ & 13.5K & 20.01 & 36.55 & 13.30\% & 19.98 & 36.49 & 13.30\% & 20.96 & 38.39 & 13.98\% & 20.14 & 36.84 & 13.37\% \\
     & SparseTSF+SA & 8.4M & 16.16 & 26.74 & 10.75\% & 18.12 & 30.96 & 12.25\% & 21.13 & 37.57 & \underline{14.40\%} & 18.13 & 31.04 & 12.13\% \\
     \cline{2-15}
     & \proposed{} & 13K & \textbf{14.79} & \textbf{24.95} & \textbf{9.32\%} & \textbf{16.64} & \textbf{28.57} & \textbf{10.67\%} & \textbf{19.02} & \textbf{33.88} & \textbf{12.68\%} & \textbf{16.48} & \textbf{28.43} & \textbf{10.66\%} \\
     \hline \hline
     \multirow{19}{*}{GBA} & HL & -- & 32.57 & 48.42 & 22.78\% & 53.79 & 77.08 & 43.01\% & 92.64 & 126.22 & 92.85\% & 56.44 & 79.82 & 48.87\% \\
     & Linear & 26K & 20.29 & 32.36 & 17.75\% & 22.61 & 36.93 & 19.85\% & 28.43 & 44.13 & 25.08\% & 22.68 & 36.59 & 20.26\% \\
     & LSTM & 98K & 20.41 & 33.47 & 15.60\% & 27.50 & 43.64 & 23.25\% & 38.85 & 60.46 & 37.47\% & 27.88 & 44.23 & 24.31\% \\
     & ASTGCN & 22.3M & 21.40 & 33.61 & 17.65\% & 26.70 & 40.75 & 24.02\% & 33.64 & 51.21 & 31.15\% & 26.15 & 40.25 & 23.29\% \\
     & DCRNN & 373K & 18.25 & 29.73 & 14.37\% & 22.25 & 35.04 & 19.82\% & 28.68 & 44.39 & 28.69\% & 22.35 & 35.26 & 20.15\% \\
     & AGCRN & 777K & 18.11 & 30.19 & \underline{13.64\%} & 20.86 & 34.42 & 16.24\% & 24.06 & 39.47 & 19.29\% & 20.55 & 33.91 & 16.06\% \\
     & STGCN & 1.3M & 20.62 & 33.81 & 15.84\% & 23.19 & 37.96 & 18.09\% & 26.53 & 43.88 & 21.77\% & 23.03 & 37.82 & 18.20\% \\
     & GWNET & 344K & 17.74 & 28.92 & 14.37\% & 20.98 & 33.50 & 17.77\% & 25.39 & 40.30 & 22.99\% & 20.78 & 33.32 & 17.76\% \\
     & STGODE & 788K & 18.80 & 30.53 & 15.67\% & 22.19 & 35.91 & 18.54\% & 26.27 & 43.07 & 22.71\% & 21.86 & 35.57 & 18.33\% \\
     & DSTAGNN & 26.9M & 19.87 & 31.54 & 16.85\% & 23.89 & 38.11 & 19.53\% & 28.48 & 44.65 & 24.65\% & 23.39 & 37.07 & 19.58\% \\
     & DGCRN & 374K & 18.09 & 29.27 & 15.32\% & 21.18 & 33.78 & 18.59\% & 25.73 & 40.88 & 23.67\% & 21.10 & 33.76 & 18.58\% \\
     & D$^2$STGNN & 446K & \underline{17.20} & \underline{28.50} & \textbf{12.22\%} & 20.80 & 33.53 & \underline{15.32\%} & 25.72 & 40.90 & 19.90\% & 20.71 & 33.44 & \underline{15.23\%} \\
     & TSMixer & 3.1M & 18.58 & 34.40 & 17.43\% & 20.78 & 37.68 & 18.74\% & 23.11 & 44.47 & 20.35\% & 20.42 & 41.52 & 18.22\% \\
     & RPMixer & 1.6M & 17.79 & 28.86 & 15.18\% & 20.07 & 32.66 & 17.50\% & 22.54 & \underline{37.22} & 20.70\% & 19.71 & \underline{32.16} & 17.27\% \\
     \cline{2-15}
     & SparseTSF & 73 & 25.41 & 44.40 & 20.66\% & 25.46 & 44.40 & 20.78\% & 26.66 & 45.52 & 23.49\% & 25.55 & 44.55 & 20.95\% \\
     & SparseTSF+MLP & 2K & 22.22 & 38.74 & 18.54\% & 22.29 & 38.74 & 18.59\% & 23.16 & 39.71 & 19.48\% & 22.36 & 38.89 & 18.64\% \\
     & SparseTSF+MLP$\times 4$ & 13.5K & 21.11 & 36.62 & 16.93\% & 21.09 & 36.58 & 16.87\% & 21.98 & 37.93 & 18.34\% & 21.23 & 36.81 & 17.12\% \\
     & SparseTSF+SA & 8.4M & 18.38 & 29.16 & 17.28\% & \underline{19.61} & \underline{32.49} & 17.19\% & \underline{21.93} & 37.41 & \underline{18.05\%} & \underline{19.42} & 32.30 & 16.62\% \\ 
     \cline{2-15}
     & \proposed{} & 13K & \textbf{16.07} & \textbf{27.57} & \textbf{12.22\%} & \textbf{17.87} & \textbf{30.83} & \textbf{13.99\%} & \textbf{20.12} & \textbf{35.37} & \textbf{16.27\%} & \textbf{17.69} & \textbf{30.61} & \textbf{13.82\%} \\
     \hline \hline
     \multirow{17}{*}{GLA} & HL & -- & 33.66 & 50.91 & 19.16\% & 56.88 & 83.54 & 34.85\% & 98.45 & 137.52 & 71.14\% & 59.58 & 86.19 & 38.76\% \\
     & Linear & 26K & 21.36 & 33.81 & 14.79\% & 23.13 & 38.70 & 16.36\% & 25.11 & 43.48 & 18.10\% & 22.85 & 37.94 & 16.50\% \\
     & LSTM & 98K & 20.09 & 32.41 & 11.82\% & 27.80 & 44.10 & 16.52\% & 39.61 & 61.57 & 25.63\% & 28.12 & 44.40 & 17.31\% \\
     & ASTGCN & 59.1M & 21.11 & 34.04 & 12.29\% & 28.65 & 44.67 & 17.79\% & 39.39 & 59.31 & 28.03\% & 28.44 & 44.13 & 18.62\% \\
     & DCRNN & 373K & 18.33 & 29.13 & 10.78\% & 22.70 & 35.55 & 13.74\% & 29.45 & 45.88 & 18.87\% & 22.73 & 35.65 & 13.97\% \\
     & AGCRN & 792K & 17.57 & 30.83 & 10.86\% & 20.79 & 36.09 & 13.11\% & 25.01 & 44.82 & 16.11\% & 20.61 & 36.23 & 12.99\% \\
     & STGCN & 2.1M & 19.87 & 34.01 & 12.58\% & 22.54 & 38.57 & 13.94\% & 26.48 & 45.61 & 16.92\% & 22.48 & 38.55 & 14.15\% \\
     & GWNET & 374K & 17.30 & 27.72 & 10.69\% & 21.22 & 33.64 & 13.48\% & 27.25 & 43.03 & 18.49\% & 21.23 & 33.68 & 13.72\% \\
     & STGODE & 841K & 18.46 & 30.05 & 11.94\% & 22.24 & 36.68 & 14.67\% & 27.14 & 45.38 & 19.12\% & 22.02 & 36.34 & 14.93\% \\
     & DSTAGNN & 66.3M & 19.35 & 30.55 & 11.33\% & 24.22 & 38.19 & 15.90\% & 30.32 & 48.37 & 23.51\% & 23.87 & 37.88 & 15.36\% \\
     & TSMixer & 4.6M & 18.20 & 82.74 & 11.56\% & 20.82 & 72.65 & 13.44\% & 23.18 & 63.78 & 15.46\% & 20.41 & 85.02 & 13.19\% \\ 
     & RPMixer & 2.6M & 17.35 & \underline{27.65} & 11.26\% & 19.97 & \underline{31.90} & 13.56\% & 22.60 & \underline{36.67} & 16.43\% & 19.60 & \underline{31.42} & 13.32\% \\ 
     \cline{2-15}
     & SparseTSF & 73 & 24.92 & 45.01 & 16.69\% & 24.95 & 45.01 & 16.74\% & 26.13 & 46.10 & 18.68\% & 25.06 & 45.17 & 16.91\% \\
     & SparseTSF+MLP & 2K & 22.47 & 40.07 & 15.21\% & 22.50 & 40.04 & 15.19\% & 23.42 & 41.14 & 16.38\% & 22.61 & 40.23 & 15.37\% \\
     & SparseTSF+MLP$\times 4$ & 13.5K & 20.80 & 36.84 & 13.52\% & 20.80 & 36.81 & 13.52\% & 21.77 & 38.49 & 14.37\% & 20.94 & 37.09 & 13.64\% \\
     & SparseTSF+SA & 8.4M & \underline{16.66} & 27.81 & \underline{10.49\%} & \underline{18.84} & 32.21 & \underline{12.37\%} & \underline{21.24} & 37.44 & \underline{14.35\%} & \underline{18.57} & 31.88 & \underline{12.27\%} \\
     \cline{2-15}
     & \proposed{} & 13K & \textbf{15.05} & \textbf{25.62} & \textbf{8.84\%} & \textbf{16.92} & \textbf{29.13} & \textbf{10.33\%} & \textbf{19.47} & \textbf{34.43} & \textbf{12.54\%} & \textbf{16.79} & \textbf{29.03} & \textbf{10.31\%} \\ 
     \hline \hline
     \multirow{14}{*}{CA} & HL & -- & 30.72 & 46.96 & 20.43\% & 51.56 & 76.48 & 37.22\% & 89.31 & 125.71 & 76.80\% & 54.10 & 78.97 & 41.61\% \\
     & Linear & 26K & 18.42 & 31.06 & 14.85\% & 21.54 & 36.26 & 18.50\% & 23.36 & 40.72 & 20.13\% & 20.96 & 35.56 & 17.42\% \\
     & LSTM & 98K & 19.01 & 31.21 & 13.57\% & 26.49 & 42.54 & 20.62\% & 38.41 & 60.42 & 31.03\% & 26.95 & 43.07 & 21.18\% \\
     & DCRNN & 373K & 17.52 & 28.18 & 12.55\% & 21.72 & 34.19 & 16.56\% & 28.45 & 44.23 & 23.57\% & 21.81 & 34.35 & 16.92\% \\
     & STGCN & 4.5M & 19.14 & 32.64 & 14.23\% & 21.65 & 36.94 & 16.09\% & 24.86 & 42.61 & 19.14\% & 21.48 & 36.69 & 16.16\% \\
     & GWNET & 469K & 16.93 & 27.53 & 13.14\% & 21.08 & 33.52 & 16.73\% & 27.37 & 42.65 & 22.50\% & 21.08 & 33.43 & 16.86\% \\
     & STGODE & 1.0M & 17.59 & 31.04 & 13.28\% & 20.92 & 36.65 & 16.23\% & 25.34 & 45.10 & 20.56\% & 20.72 & 36.65 & 16.19\% \\
     & TSMixer & 9.5M & 17.62 & 64.15 & 14.49\% & 19.62 & 44.03 & 15.59\% & 22.09 & 46.29 & 18.32\% & 19.19 & 48.99 & 15.36\% \\
     & RPMixer & 7.1M & 16.36 & 26.39 & 13.02\% & 18.21 & \underline{29.75} & \underline{14.23\%} & 20.49 & \underline{34.11} & 16.19\% & 17.95 & \underline{29.42} & 14.18\% \\
     \cline{2-15}
     & SparseTSF & 73 & 23.19 & 42.32 & 17.80\% & 23.21 & 42.31 & 17.84\% & 24.50 & 43.45 & 20.39\% & 23.36 & 42.48 & 18.12\% \\
     & SparseTSF+MLP & 2K & 20.77 & 37.26 & 16.26\% & 20.75 & 37.21 & 16.22\% & 22.04 & 38.52 & 17.85\% & 20.94 & 37.44 & 16.47\% \\
     & SparseTSF+MLP$\times 4$ & 13.5K & 18.95 & 33.69 & 14.33\% & 18.93 & 33.64 & 14.31\% & 19.96 & 35.54 & 15.25\% & 19.09 & 33.96 & 14.45\% \\
     & SparseTSF+SA & 8.4M & \underline{15.64} & \underline{26.18} & \underline{11.68\%} & \underline{17.78} & 30.15 & 15.06\% & \underline{20.02} & 35.34 & \underline{15.92\%} & \underline{17.39} & 29.77 & \underline{13.75\%} \\
     \cline{2-15}
     & \proposed{} & 13K & \textbf{14.15} & \textbf{24.44} & \textbf{10.01\%} & \textbf{15.89} & \textbf{27.70} & \textbf{11.55\%} & \textbf{18.16} & \textbf{32.50} & \textbf{13.74\%} & \textbf{15.75} & \textbf{27.58} & \textbf{11.50\%} \\
     \shline
\end{tabular}%
}
\end{center}
\vskip -0.1in
\end{table*}

\subsection{Benchmark Result}
\label{sec:benchmark} 
We used the same hyperparameter setting ($n_\text{block}=4$, $d=16$, $w=12$) for the proposed \proposed{} method across all four datasets.
This hyperparameter configuration was determined based on the study presented in \cref{sec:hyper_parameter}.
The benchmark results are presented in \cref{tab:largest_benchmark}.

Comparing the proposed \proposed{} method with other methods, we observed that it achieved the best performance across all four datasets.
The proposed method has attained state-of-the-art level performance.
Moreover, when considering the model size (i.e., number of learnable parameters), it is also relatively lightweight.
The second-best method across different datasets uses at least 10 times more parameters than the proposed method.

Next, we focus on the performance of SparseTSF and its variants, as these methods serve not only as baseline methods but also for ablation studies.
First, SparseTSF shows much worse performance compared to the proposed method, indicating significant room for improvement.
The first attempted improvement utilizes MLP instead of a linear layer in the cross-period sparse forecast component.
We call this variant SparseTSF+MLP in the table.
This improvement, proposed by~\cite{lin2024sparsetsf}, enhances the SparseTSF method, but its performance still falls short of the proposed method.

We then increased the number of layers in the MLP to match the parameter count of the proposed method.
We call this variant SparseTSF+MLP$\times 4$ in the table.
Increasing the layer count further improves performance, allowing for a fair comparison between systems with and without the proposed ultra-compact shape bank component.
When comparing these two methods, the proposed method again outperforms the alternative, demonstrating the necessity of adopting the proposed ultra-compact shape bank component in the forecasting system.

Lastly, SparseTSF+SA also uses a shape bank component, but instead of the proposed ultra-compact shape bank design, it employs a standard attention design~\cite{vaswani2017attention}.
SparseTSF+SA improves upon the SparseTSF+MLP$\times 4$ method, showing the importance of modeling intra-period dependency.
However, the standard attention design also increases the cost of deploying the model as it dramatically increases the model size.
Comparing SparseTSF+SA with the proposed \proposed{}, we observe that the proposed method is not only much lighter but also performs better.
This observation further reinforces the necessity of the proposed ultra-compact design over the standard design.

In summary, comparison with SparseTSF+MLP$\times 4$ demonstrates the importance of modeling intra-period dependency.
Comparison with SparseTSF+SA shows that the proposed ultra-compact shape bank design is both more efficient and effective than the standard design.
Finally, the proposed \proposed{} is the best method, achieving superior performance while being at least 10 times lighter than the second-best methods.

\subsection{Generalization Capability}
\label{sec:generalization}
To evaluate the generalization capability of different models across regions, we trained the models on one dataset and tested them on another.
We excluded the CA dataset from this set of experiments as it is a superset of the other three datasets.
Only methods capable of handling dimension mismatches between the training and test datasets were included in this study.
These comprise methods applied through the channel independent strategy (i.e., Linear, SparseTSF, and the proposed \proposed{}) and a method that learns dependencies between different dimensions using an attention mechanism (i.e., iTransformer).
The results of the generalization capability study are reported in \cref{tab:generalization}.

\begin{table}[htp]
\caption{
Generalization capability of models across different regions.
Performances are reported as average MAE.
Due to scalability issues with larger datasets, iTransformer was only tested using the SD dataset for training.
}
\label{tab:generalization}
\begin{center}
\resizebox{0.85\linewidth}{!}{%
\begin{tabular}{llcccc}
    \shline
    Train & \multirow{2}{*}{Method} & \multirow{2}{*}{Param} & \multicolumn{3}{c}{Test Data} \\ \cline{4-6} 
    Data&  &  & SD & GBA & GLA \\ 
     \hline \hline
     \multirow{7}{*}{SD} & Linear & 26K & 23.01 & 52.66 & 53.45 \\
     & iTransformer & 6.5M & 19.63 & 108.10 & 110.56 \\
     & SparseTSF & 73 & 24.66 & 28.03 & 27.36 \\
     & SparseTSF+MLP & 2K & 21.85 & 23.66 & 23.66 \\
     & SparseTSF+MLPx4 & 13.5K & 20.14 & 21.84 & 21.87 \\
     & SparseTSF+SA & 8.4M & 18.13 & 35.31 & 37.80 \\
     \cline{2-6}
     & \proposed{} & 13K & \textbf{16.48} & \textbf{18.41} & \textbf{18.01} \\
     \hline \hline
     \multirow{6}{*}{GBA} & Linear & 26K & 73.40 & 22.68 & 67.92 \\
     & SparseTSF & 73 & 25.33 & 25.55 & 25.68 \\
     & SparseTSF+MLP & 2K & 22.16 & 22.36 & 22.88 \\
     & SparseTSF+MLPx4 & 13.5K & 20.84 & 21.23 & 21.33 \\
     & SparseTSF+SA & 8.4M & 18.68 & 19.42 & 19.49 \\
     \cline{2-6}
     & \proposed{} & 13K & \textbf{16.83} & \textbf{17.69} & \textbf{17.59} \\
     \hline \hline
     \multirow{6}{*}{GLA} & Linear & 26K & 22.81 & 23.43 & 22.85 \\
     & SparseTSF & 73 & 24.74 & 25.88 & 25.06 \\
     & SparseTSF+MLP & 2K & 22.26 & 23.36 & 22.61 \\
     & SparseTSF+MLPx4 & 13.5K & 20.77 & 21.72 & 20.94 \\
     & SparseTSF+SA & 8.4M & 32.37 & 29.65 & 18.57 \\
     \cline{2-6}
     & \proposed{} & 13K & \textbf{16.46} & \textbf{17.68} & \textbf{16.79} \\
     \shline
\end{tabular}%
}
\end{center}
\vskip -0.1in
\end{table}

Methods that leverage the periodicity of time series data, such as SparseTSF and \proposed{}, typically demonstrate good generalizability.
The performance difference between results obtained from the test set of the same region and a different region is minimal for these methods.
One possible reason for this enhanced generalizability could be that the periodicity assumption serves as a generally effective inductive bias for building time series models.
Conversely, both the linear model and iTransformer, which do not utilize the periodicity assumption, cannot generalize effectively across datasets.
Their performance degrades significantly when tested on datasets from different regions.

\subsection{Scaling Capability}
\label{sec:scaling}
\begin{figure}[htp]
\begin{center}
\centerline{
\includegraphics[width=0.99\linewidth]{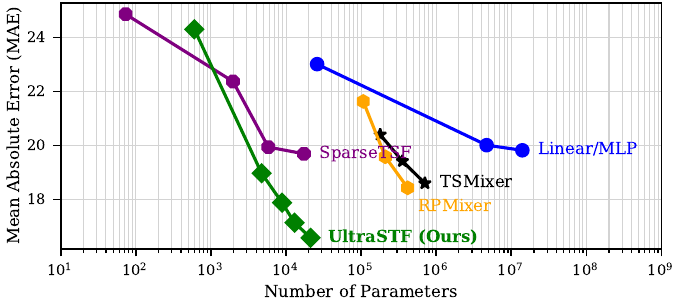}
}
\caption{
Scaling capabilities of different methods evaluated on SD's test data.
The proposed \proposed{} demonstrates superior parameter efficiency.
}
\label{fig:scaling_test_sd}
\end{center}
\end{figure}

This section evaluates and compares the scaling capabilities of different models.
We increase the model size by increasing the number of layers for each method and observed the corresponding changes in performance.
\cref{fig:scaling_test_sd} presents the experimental results for the SD dataset.
The proposed \proposed{} method not only achieves the best overall performance but also demonstrates superior scaling capability compared to other methods.
As the model size increases, \proposed{} exhibits more consistent and substantial improvements in performance, indicating its effective utilization of additional parameters.
This enhanced scaling capability suggests that \proposed{} is well-suited for handling larger and more complex datasets, potentially offering even greater advantages as computational resources expand.

To ensure the robustness of our findings, we conduct the same scaling analysis using validation data.
The results, presented in \cref{app:scaling}, corroborate our observations from the test set, further validating the superior scaling capability of \proposed{}.
This consistency across both test and validation sets strengthens the reliability of our conclusions and underscores the generalizability of \proposed{}'s performance advantages.

\subsection{Hyper-Parameter Study}
\label{sec:hyper_parameter}
\begin{figure}[htp]
\begin{center}
\centerline{
\includegraphics[width=0.99\linewidth]{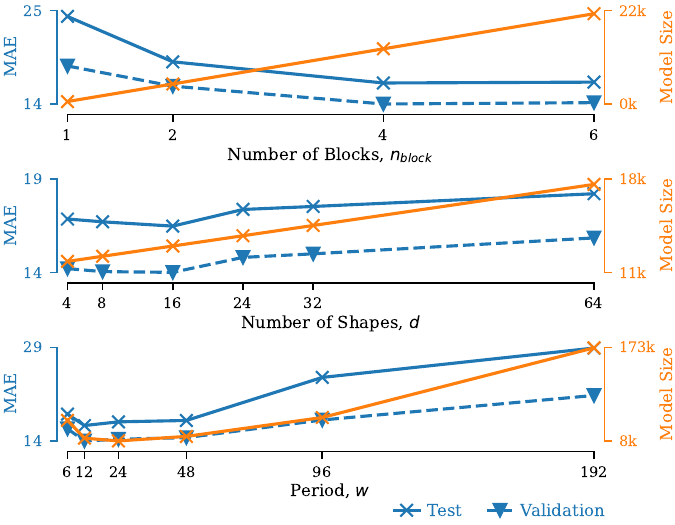}
}
\caption{
The effect of different hyper-parameter settings on model performance (i.e., average MAE) and size. 
We have shown the MAE on both the test and validation partitions of the SD dataset.
}
\label{fig:param_study}
\end{center}
\end{figure}

We use the SD dataset to study the effect of different hyper-parameters on our proposed model architecture.
We focus on three key hyper-parameters: the number of core blocks ($n_\text{block}$), the number of shapes ($d$) in the shape bank, and the period ($w$).
Our analysis examines these hyper-parameters' impact on both forecasting accuracy (measured by average MAE) and model size.
The experimental results for this hyper-parameter study are presented in \cref{fig:param_study}.
We report the average MAE on both test and validation partitions to assess the generalizability of the hyper-parameters across dataset partitions.

The number of core blocks~$n_\text{block}$ demonstrates a linear relationship with the model size, consistent with our analysis in \cref{sec:complexity}.
As we increase the number of core blocks, we observe a reduction in MAE.
However, the improvement in MAE becomes minimal between $n_\text{block}=4$ and $n_\text{block}=6$ for both validation and test sets.
Consequently, we set $n_\text{block}=4$ in all our subsequent experiments to balance performance and model complexity.
The number of shapes~$d$ in the shape bank also exhibits a linear relationship with model size, aligning with our analysis in \cref{sec:complexity}.
Regarding MAE, we find that setting $d=16$ yields the best performance across both validation and test sets.
Therefore, we adopt $d=16$ for all our other experiments.
The relationship between the period~$w$ and model size is more complex.
Among the six $w$ settings tested, $w=24$ produces the smallest model, with size increasing for both smaller and larger values.
Considering MAE, we observe that $w=12$ leads to the best performance on both test and validation sets.
As a result, we set $w=12$ in all our other experiments.

\subsection{Shape Bank Visualization}
\label{sec:visualization}
\begin{figure}[htp]
\begin{center}
\centerline{
\includegraphics[width=0.95\linewidth]{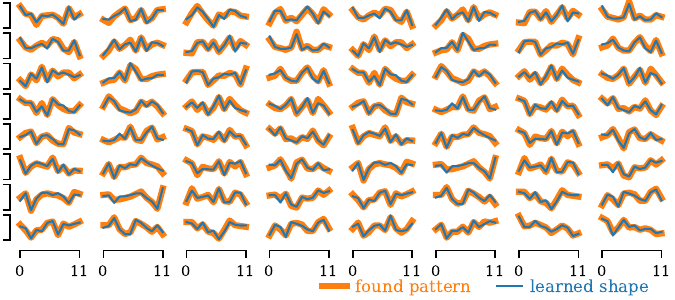}
}
\caption{
The shape learned by the model can be found in the dataset.
}
\label{fig:learned_shape_sd}
\end{center}
\end{figure}

Our model, with the hyperparameters found in \cref{sec:hyper_parameter}, learns 64 shapes through its shape bank components.
To validate the relevance of these patterns, we conducted an analysis using the SD dataset.
We extracted the 64 shapes from the trained SD model and searched for their closest matches within the SD dataset using Matrix Profile~\cite{yeh2016matrix}.
\cref{fig:learned_shape_sd} presents these 64 learned shapes alongside their corresponding matches from the dataset.
Visual inspection confirms that each shape indeed represents a genuine pattern found in the original data, demonstrating the model's ability to capture meaningful temporal structures.
For additional visualization results, please refer to \cref{app:visualization}.

\subsection{Sensitivity Analysis}
\label{sec:sensitivity}
\begin{figure}[htp]
\begin{center}
\includegraphics[width=0.99\linewidth]{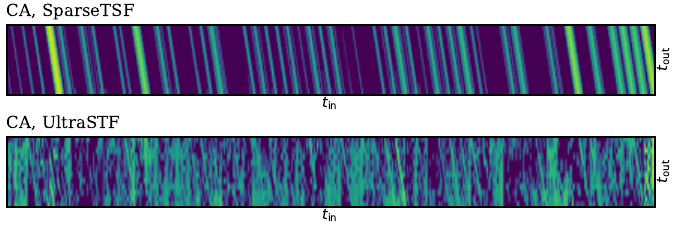}
\caption{
Sensitivity analysis of models trained on the CA datasets.
The proposed \proposed{} method learns both cross-period and intra-period relationships, while the SparseTSF method primarily captures cross-period relationships.
}
\label{fig:viz_model}
\end{center}
\end{figure}

To verify that the proposed \proposed{} method indeed learns both intra-period and cross-period relationships, we conducted a sensitivity analysis on both \proposed{} and SparseTSF methods.
We focus on comparing with SparseTSF because it predominantly models cross-period relationships.
The analysis was performed by plotting the output of each model when fed a $t_\text{in}$ by $t_\text{in}$ identity matrix, allowing us to visualize each model's response when the input ``1" is at different time steps.
\Cref{fig:viz_model} shows the results of this analysis for model trained on CA datasets.

When examining the figures associated with SparseTSF, we observe many prominent diagonal lines, each spanning a period length, indicating that the model predominantly learns cross-period relationships.
In contrast, the figures associated with \proposed{} reveal more complex patterns, suggesting that the proposed method indeed models both intra-period and cross-period relationships.
This sensitivity analysis provides visual evidence of the fundamental difference between \proposed{} and SparseTSF in their ability to capture temporal dependencies, supporting our claim that \proposed{} offers a more comprehensive modeling approach for spatio-temporal forecasting tasks.
\section{Conclusion}
In this paper, we have proposed \proposed{}, a novel method that combines an ultra-compact shape bank component with a cross-period sparse forecast component.
\proposed{} effectively leverages the inherent periodicity typically found in spatio-temporal data, modeling intra-period dependencies through the shape bank component and inter-period dependencies via the cross-period sparse forecast component.
Our extensive experiments demonstrate that \proposed{} achieves state-of-the-art performance while maintaining a lightweight architecture compared to alternative solutions.
Notably, \proposed{} exhibits remarkable generalization capabilities across different datasets, highlighting its potential as an effective model architecture for a time series foundation model~\cite{yeh2023toward}.
The lightweight nature of \proposed{} makes it an attractive candidate for edge computing applications and real-time forecasting scenarios.

\bibliographystyle{ACM-Reference-Format}
\bibliography{section/reference}

\clearpage
\appendix

\section{Implementation Detail}
\label{app:implementation}
This section provides a detailed overview of the PyTorch implementation of the proposed \proposed{} method.
The main class of the \proposed{} method is presented in \cref{alg:main_class}.

\begin{algorithm}[htp]
\caption{The main class of \proposed{}}\label{alg:main_class}
\begin{minted}[breaklines,xleftmargin=2em,linenos,fontsize=\scriptsize, escapeinside=!!]{python}
import torch
import torch.nn as nn
from collections import OrderedDict

class UltraSTF(nn.Module):
  def __init__(self, seq_len, pred_len,  
         period_len, n_shape, n_layer):
    super(UltraSTF, self).__init__()

    self.aggregation = Aggregation(
      seq_len, period_len) # !\cref{alg:aggregation}!

    layers = OrderedDict()
    for i in range(n_layer):
      if i == n_layer - 1:
        pred_len_ = pred_len
      else:
        pred_len_ = seq_len
      layers[f'block_{i}'] = Block(
        seq_len, pred_len_,
        period_len, n_shape) # !\cref{alg:block}!
    block = nn.Sequential(layers)
    self.add_module('block', block)
    self.block = block

  def forward(self, x):
    # x = [batch_size, seq_dim, seq_len, ]
    # y = [batch_size, seq_dim, pred_len, ]
    x_ = x.detach()
    x_mu = torch.mean(x_, 2, keepdim=True)
    x_sigma = torch.std(x_, 2, keepdim=True)
    x_sigma[x_sigma < 1e-6] = 1.0
    x = (x - x_mu) / x_sigma

    h = self.aggregation(x)
    y = self.block(h)

    y = (y * x_sigma) + x_mu
    return y
\end{minted}
\end{algorithm}

The implementation process begins with input data normalization using mean and standard deviation.
Next, the data is processed through the aggregation module, as detailed in \cref{alg:aggregation}.
The data is then processed sequentially through multiple core blocks, with the detailed algorithm for each block presented in \cref{alg:block}.
Finally, the previously removed mean and standard deviation are reapplied to the output.

The aggregation module, shown in \cref{alg:aggregation}, utilizes a single layer of 1D convolutional operation.
By reshaping the input data, the 1D convolutional filter is applied independently to all channels.
This implementation corresponds to the operation depicted in \cref{fig:sparsetsf}.\textit{bottom left}.

\begin{algorithm}[htp]
\caption{The aggregation class of \proposed{}}\label{alg:aggregation}
\begin{minted}[breaklines,xleftmargin=2em,linenos,fontsize=\scriptsize, escapeinside=!!]{python}
import torch.nn as nn

class Aggregation(nn.Module):
  def __init__(self, seq_len, period_len):
    super(Aggregation, self).__init__()
    padding = int(period_len // 2)
    kernel_size = int(1 + 2 * padding)
    self.conv1d = nn.Conv1d(
      in_channels=1, out_channels=1,
      kernel_size=kernel_size,
      stride=1, padding=padding,
      padding_mode="zeros", bias=False)
    self.seq_len = seq_len

  def forward(self, x):
    seq_dim = x.size(1)
    h = x.reshape(-1, 1, self.seq_len)
    h = self.conv1d(h)
    h = h.reshape(-1, seq_dim, self.seq_len)
    return h + x
\end{minted}
\end{algorithm}

\sloppy
The core block module, detailed in \cref{alg:block}, consists of two main components: the intra-period shape bank component and the cross-period sparse forecasting component.
In the initialization function, we set up \texttt{intra\_key}, \texttt{intra\_value}, and \texttt{linear\_intra\_query} for the shape bank component, and \texttt{linear\_inter} for the cross-period sparse forecasting component.

\begin{algorithm}[htp]
\caption{The core block class of \proposed{}}\label{alg:block}
\begin{minted}[breaklines,xleftmargin=2em,linenos,fontsize=\scriptsize, escapeinside=!!]{python}
import numpy as np
import torch
import torch.nn as nn

class Block(nn.Module):
  def __init__(self, seq_len, pred_len,  
         period_len, n_shape):
    super(Block, self).__init__()
    n_seg_x = int(np.floor(seq_len / period_len))
    n_seg_y = int(np.ceil(pred_len / period_len))

    intra_key = torch.randn(n_shape, period_len)
    self.intra_key = nn.Parameter(intra_key)

    intra_value = torch.randn(n_shape, period_len)
    self.intra_value = nn.Parameter(intra_value)

    self.linear_intra_query = nn.Linear(
      period_len, period_len, bias=False)
    self.linear_inter = nn.Linear(
      n_seg_x, n_seg_y, bias=False)

    self.pred_len = pred_len
    self.period_len = period_len
    self.n_seg_x = n_seg_x
    self.n_seg_y = n_seg_y

  def forward(self, x):
    seq_len_norm = self.n_seg_x * self.period_len
    pred_len_norm = self.n_seg_y * self.period_len
    seq_dim = x.size(1)
    x = x[:, :, :seq_len_norm]
    x = x.reshape(
      -1, seq_dim, self.n_seg_x,
      self.period_len)

    query = self.linear_intra_query(x)
    key = self.intra_key
    value = self.intra_value

    key = key.permute(1, 0)
    attn = torch.matmul(query, key)
    attn = nn.ReLU()(attn)
    h = torch.matmul(attn, value)
    x = x + h

    x = x.reshape(
      -1, self.n_seg_x, self.period_len)
    x = x.permute(0, 2, 1)

    y = self.linear_inter(x)
    y = y.permute(0, 2, 1)
    y = y.reshape(-1, seq_dim, pred_len_norm)
    y = y[:, :, :self.pred_len]
    return y
\end{minted}
\end{algorithm}

For the shape bank component, we use the \texttt{linear\_intra\_query} to process the input to obtain the query.
Next, we compute the attention score with \texttt{intra\_key}, and multiply the score with \texttt{intra\_value} to obtain the output.
To extend the shape bank mechanism from single head to multi-head, we need to first replace lines 18 and 19 in \cref{alg:block} with \cref{alg:mh_init} where the weights associated with the multi-head attention mechanism are initialized.
Next, we also need to replace the code from lines 37 to 44 in \cref{alg:block} with \cref{alg:mh_forward} which carries out the multi-head attention mechanism.

\begin{algorithm}[htp]
\caption{Initialization for multi-head attention}\label{alg:mh_init}
\begin{minted}[breaklines,xleftmargin=2em,linenos,fontsize=\scriptsize, escapeinside=!!]{python}
self.linear_intra_query = []
self.linear_intra_key = []
self.linear_intra_value = []
for i in range(n_head):
  self.linear_intra_query.append(nn.Linear(
    period_len, period_len // n_head, bias=False))
  self.linear_intra_key.append(nn.Linear(
    period_len, period_len // n_head, bias=False))
  self.linear_intra_value.append(nn.Linear(
    period_len, period_len // n_head, bias=False))
  self.register_module(f'linear_intra_query_{i}', 
                       self.linear_intra_query[i])
  self.register_module(f'linear_intra_key_{i}', 
                       self.linear_intra_key[i])
  self.register_module(f'linear_intra_value_{i}', 
                       self.linear_intra_value[i])
self.linear_intra = nn.Linear(
  (period_len // n_head) * n_head, period_len, 
  bias=False)
\end{minted}
\end{algorithm}

\begin{algorithm}[htp]
\caption{Forward pass for multi-head attention}\label{alg:mh_forward}
\begin{minted}[breaklines,xleftmargin=2em,linenos,fontsize=\scriptsize, escapeinside=!!]{python}
query = x
key = self.intra_key
value = self.intra_value
h = []
for i in range(self.n_head):
  query_i = self.linear_intra_query[i](query)
  key_i = self.linear_intra_key[i](key)
  value_i = self.linear_intra_value[i](value)
  key_i = key_i.permute(1, 0)
  attn_i = torch.matmul(query_i, key_i)
  attn_i = nn.ReLU()(attn_i)
  h.append(torch.matmul(attn_i, value_i))
h = torch.concat(h, 3)
h = self.linear_intra(h)
\end{minted}
\end{algorithm}

After the shape bank component, we use the \texttt{linear\_inter} layer for the cross-period sparse forecasting mechanism.
It's worth noting that our implementation can handle cases where $\frac{t_\text{in}}{w}$ and/or $\frac{t_\text{out}}{w}$ are not integers.

\section{Detailed Complexity Analysis}
\label{app:complexity}
We analyze the time complexity of our model by examining each process illustrated in \cref{fig:overall}.
The convolutional layer's time complexity is $O(n t_\text{in})$.
For the shape bank component, the linear layer has a time complexity of $O(n w^2 k_\text{in}) = O(n w t_\text{in})$.
Since $w$ is a constant, this simplifies to $O(n t_\text{in})$.
Both multiplications in the shape bank component have a time complexity of $O(n w d k_\text{in}) = O(n d t_\text{in})$.
As $d$ is also a constant, this again simplifies to $O(n t_\text{in})$.
The cross-period sparse forecast component's time complexity is either $O(n w k_\text{in}^2) = O(n \frac{t_\text{in}^2}{w^2})= O(n t_\text{in}^2)$ or $O(n w k_\text{in} k_\text{out}) = O(n \frac{t_\text{in} t_\text{out}}{w^2})=O(n t_\text{in} t_\text{out})$.
For spatio-temporal datasets, the typical relationship between $n$, $t_\text{in}$, and $t_\text{out}$ is $n > t_\text{in} \geq t_\text{out}$.
Therefore, the overall time complexity of our proposed method is $O(n t_\text{in}^2)$.
This complexity is comparable to that of SparseTSF, which also has a time complexity of $O(n t_\text{in}^2)$, indicating that our method is suitable for large-scale datasets.

The space complexity of our proposed method can be derived from \cref{fig:overall}.
We begin by examining the sizes of the input, intermediate, and output tensors.
Four distinct tensor sizes are identified: 1) $n \times t_\text{in}$, 2) $n \times w \times k_\text{in}$, 3) $n \times t_\text{out}$, and 4) $n \times w \times k_\text{out}$.
Given that $t_\text{in} = w \times k_\text{in}$ and $t_\text{out} = w \times k_\text{out}$, these tensors contribute two different space costs: $O(n t_\text{in})$ and $O(n t_\text{out})$.
Considering the weight tensors associated with different components, only the weight matrix from the linear layer of the cross-period sparse forecast component scales with the input time series size.
The space complexity of this weight matrix is $O(t_\text{in} t_\text{out})$ or $O(t_\text{in}^2)$.
For spatio-temporal datasets, the typical relationship between $n$, $t_\text{in}$, and $t_\text{out}$ is $n > t_\text{in} \geq t_\text{out}$.
Consequently, the overall space complexity of our proposed method is $O(n t_\text{in})$.
This space complexity is equivalent to that of the SparseTSF baseline method, indicating that our approach is suitable for large-scale datasets.

The number of parameters within the proposed method can be computed with the following equation.

\vspace{-1em}

\begin{align*}
\floor*{\frac{w}{2}} + 1 & + (n_\text{block} - 1) \left(w^2 + 2wd + \frac{t_\text{in}^2}{w^2} \right) \\
& + \left(w^2 + 2wd + \frac{t_\text{in} t_\text{out}}{w^2} \right) \\
\end{align*}

\vspace{-1em}

\noindent The first term, $\floor*{\frac{w}{2}} + 1$, is from the aggregation module, i.e., the filter of the convolutional layer.
In each block, the term $w^2 + 2wd$ is from the proposed ultra-compact shape bank, and the terms $\frac{t_\text{in}^2}{w^2}$ and $\frac{t_\text{in} t_\text{out}}{w^2}$ are from the cross-period sparse forecast module.
Comparing to the SparseTSF model, the extra cost in terms of learnable parameters is the $w^2 + 2wd$ term from each block. 
This extra cost is negligible, as both $w$ and $d$ are small constants and are not data-dependent.

\section{Additional Scaling Capability Result}
\label{app:scaling}
\begin{figure}[htp]
\begin{center}
\centerline{
\includegraphics[width=0.9\linewidth]{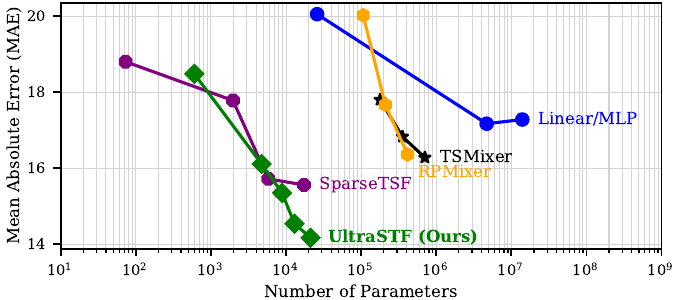}
}
\caption{
Scaling capabilities of different methods evaluated on SD's validation data.
}
\label{fig:scaling_validate_sd}
\end{center}
\end{figure}

\begin{figure}[htp]
\begin{center}
\centerline{
\includegraphics[width=0.9\linewidth]{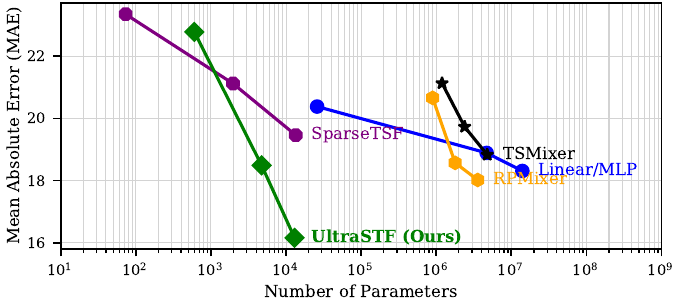}
}
\caption{
Scaling capabilities of different methods evaluated on CA's test data.
}
\label{fig:scaling_test_ca}
\end{center}
\end{figure}

\begin{figure}[htp]
\begin{center}
\centerline{
\includegraphics[width=0.9\linewidth]{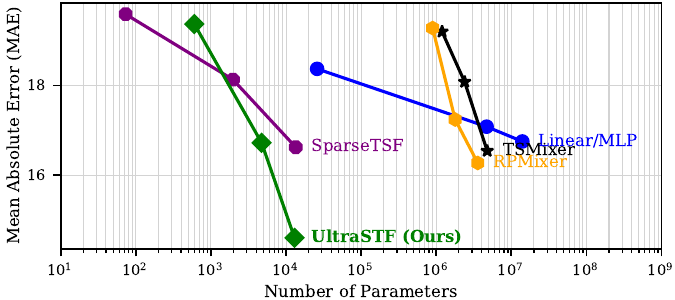}
}
\caption{
Scaling capabilities of different methods evaluated on CA's validation data.
}
\label{fig:scaling_validate_ca}
\end{center}
\end{figure}

In \cref{fig:scaling_test_sd}, we demonstrated how different models' performance improves with increasing model size using SD test data.
We extended this analysis to include SD validation and CA test and validation sets, as shown in \cref{fig:scaling_validate_sd,fig:scaling_test_ca,fig:scaling_validate_ca}.
Across all datasets and partitions, \proposed{} consistently exhibits superior scaling capability, achieving higher performance improvements as model size increases compared to competing methods.
This comprehensive evaluation provides strong evidence for \proposed{}'s superior scaling capability and parameter efficiency across different datasets.
The consistency between test and validation results further underscores the reliability of our scaling analysis, indicating that these trends are fundamental characteristics of \proposed{} rather than dataset-specific artifacts.

\section{Additional Shape Bank Visualization}
\label{app:visualization}
\begin{figure}[htp]
\begin{center}
\centerline{
\includegraphics[width=0.9\linewidth]{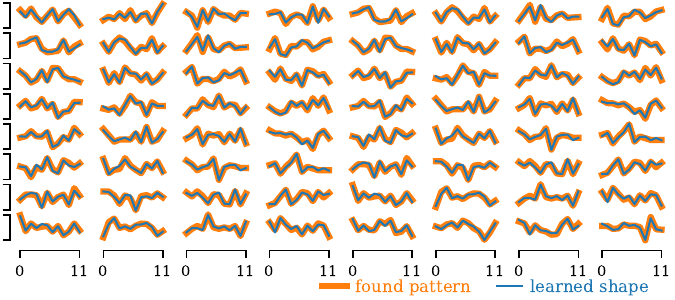}
}
\caption{
The shape learned by the model can be found in the dataset.
}
\label{fig:learned_shape_ca}
\end{center}
\end{figure}

While the example provided in \cref{sec:visualization} focused on the SD dataset, 
we aim to demonstrate that the observed phenomenon is a general occurrence.
To this end, we replicated the same experiment using the CA dataset.
The results are presented in \cref{fig:learned_shape_ca}.
Visual inspection of these results reaffirms our findings:
each shape learned by the shape bank mechanism represents a genuine pattern found in the original data.
This further demonstrates the model's ability to capture meaningful temporal structures across different datasets.

\section{Long-Term Time Series Forecasting}
\label{app:lts_forecasting}
We conducted long-term time series forecasting experiments following the methodology in~\cite{lin2024sparsetsf} to compare the proposed \proposed{} method with other time series forecasting approaches.
The experiments were performed on ETTh1, ETTh2, Electricity, and Traffic datasets, using a look-back window of 720 time steps.
Performance was measured using mean square error (MSE) for prediction horizons of 96, 192, 336, and 720 time steps.
The experimental results are presented in \cref{tab:ltsf_benchmark}.
The proposed \proposed{} method achieves comparable performance to other state-of-the-art methods.

\begin{table}[htp]
\caption{Performance comparisons of different methods on multivariate long-term time series forecasting benchmarks with various prediction horizons (96, 192, 336, and 720 time steps).
We \textbf{bold} the best-performing results and \underline{underline} the second-best results.}
\label{tab:ltsf_benchmark}
\begin{center}
\resizebox{0.8\linewidth}{!}{%
\begin{tabular}{llcccc}
    \shline
    \multirow{2}{*}{Data} & \multirow{2}{*}{Method} & \multicolumn{4}{c}{MSE} \\ \cline{3-6} 
     &  &  96 & 192 & 336 & 720 \\ 
     \hline \hline
     \multirow{7}{*}{ETTh1} & FEDformer & 0.375 & 0.427 & 0.459 & 0.484 \\
     & TimesNet & 0.384 & 0.436 & 0.491 & 0.521 \\
     & PatchTST & 0.385 & \underline{0.413} & 0.440 & 0.456 \\
     & DLinear & 0.384 & 0.443 & 0.446 & 0.504 \\
     & FITS & 0.382 & 0.417 & \underline{0.436} & \underline{0.433} \\
     & SparseTSF & \underline{0.362} & \textbf{0.403} & \textbf{0.434} & \textbf{0.426} \\
     \cline{2-6}
     & \proposed{} & \textbf{0.360} & \textbf{0.403} & 0.442 & 0.477 \\
     \hline \hline
     \multirow{7}{*}{ETTh2} & FEDformer & 0.340 & 0.433 & 0.508 & 0.480 \\
     & TimesNet & 0.340 & 0.402 & 0.452 & 0.462 \\
     & PatchTST & \underline{0.274} & \underline{0.338} & 0.367 & 0.391 \\
     & DLinear & 0.282 & 0.350 & 0.414 & 0.588 \\
     & FITS & \textbf{0.272} & \textbf{0.333} & \textbf{0.355} & \textbf{0.378} \\
     & SparseTSF & 0.294 & 0.339 & \underline{0.359} & \underline{0.383} \\
     \cline{2-6}
     & \proposed{} & 0.277 & 0.345 & 0.363 & 0.398 \\
     \hline \hline
     \multirow{7}{*}{Electricity} & FEDformer & 0.188 & 0.197 & 0.212 & 0.244 \\
     & TimesNet & 0.168 & 0.184 & 0.198 & 0.220 \\
     & PatchTST & \textbf{0.129} & \underline{0.149} & \textbf{0.166} & 0.210 \\
     & DLinear & 0.140 & 0.153 & \underline{0.169} & \textbf{0.204} \\
     & FITS & 0.145 & 0.159 & 0.175 & 0.212 \\
     & SparseTSF & 0.138 & 0.151 & \textbf{0.166} & \underline{0.205} \\
     \cline{2-6}
     & \proposed{} & \underline{0.131} & \textbf{0.148} & \textbf{0.166} & \underline{0.205} \\
     \hline \hline
     \multirow{7}{*}{Traffic} & FEDformer & 0.573 & 0.611 & 0.621 & 0.630 \\
     & TimesNet & 0.593 & 0.617 & 0.629 & 0.640 \\
     & PatchTST & \textbf{0.366} & \textbf{0.388} & \textbf{0.398} & 0.457 \\
     & DLinear & 0.413 & 0.423 & 0.437 & 0.466 \\
     & FITS & 0.398 & 0.409 & 0.421 & 0.457 \\
     & SparseTSF & 0.389 & 0.398 & \underline{0.411} & \underline{0.448} \\
     \cline{2-6}
     & \proposed{} & \underline{0.370} & \underline{0.389} & 0.427 & \textbf{0.439} \\
     \shline
\end{tabular}%
}
\end{center}
\vskip -0.1in
\end{table}

One possible explanation for the similar performance across methods is the relatively smaller dataset sizes used in long-term time series forecasting compared to spatio-temporal forecasting.
The long-term time series forecasting datasets range in size from 121,940 to 15,122,928 data points, while the spatio-temporal forecasting datasets contain between 25,088,640 and 301,344,000 data points.
Dataset size is measured in terms of the total number of numerical values in the dataset.
Consequently, when different methods reach a certain level of effectiveness on these smaller datasets, their performance results tend to converge.

\end{document}